%% file: main.tex
\documentclass[lettersize,journal]{IEEEtran}
\usepackage{amsmath,amsfonts}
\usepackage{algorithmic}
\usepackage{array}
\usepackage[caption=false,font=normalsize,labelfont=sf,textfont=sf]{subfig}
\usepackage{textcomp}
\usepackage{stfloats}
\usepackage{url}
\usepackage{verbatim}
\usepackage{graphicx}
\usepackage{enumitem}
\usepackage{amssymb}% to access $\blacksquare$
\usepackage{cite}
\usepackage{xcolor}
\newcounter{boxlblcounter}  % Creating a custom list using the list environment
\newenvironment{boxlabel}
  {\begin{list}
    {\arabic{boxlblcounter}}
    {\usecounter{boxlblcounter}
     \setlength{\labelwidth}{2em}
     \setlength{\labelsep}{0em}
     \setlength{\itemsep}{2pt}
     \setlength{\leftmargin}{0.8cm}
     \setlength{\rightmargin}{0cm}
     \setlength{\itemindent}{0em} 
     
    }
  }
{\end{list}}

\def\BibTeX{{\rm B\kern-.05em{\sc i\kern-.025em b}\kern-.08em
    T\kern-.1667em\lower.7ex\hbox{E}\kern-.125emX}}
\usepackage{balance}
\newcommand\Title{Vehicle Teleoperation: Performance Assessment of SRPT Approach Under State Estimation Errors}

\newcommand{\RN}[1]{%
  \textup{\uppercase\expandafter{\romannumeral#1}}%
}
\begin{document}

\begin{center}
\textbf{This work has been submitted to Elsevier for possible publication. Copyright may be transferred without notice, after which this version may no longer be accessible.}
\end{center}

\title{\Title}
%\author{Jai~Prakash \IEEEmembership{Fellow, IEEE},~Michele~Vignati,~Edoardo~Sabbioni,~and~Federico~Cheli% <-this % stops a space
%\thanks{The authors belong to the Department of Mechanical Engineering, Politecnico Di Milano, Italy (e-mail:
%jai.prakash@polimi.it; michele.vignati@polimi.it; edoardo.sabbioni@polimi.it; federico.cheli@polimi.it)
%}%
%}

\author{Jai~Prakash,\,
~Michele~Vignati,\,
~and~Edoardo~Sabbioni\
\thanks{The authors belong to the Department of Mechanical Engineering, Politecnico Di Milano, Italy (e-mail:
jai.prakash@polimi.it; michele.vignati@polimi.it; edoardo.sabbioni@polimi.it)
}%
}
%\markboth{IEEE/ASME Transactions on Mechatronics,~Vol.~XX, No.~X, March~2023}%
\markboth{}%
{\Title}

\maketitle
\input{sections/00-abstract}
\input{sections/01-introduction}
\input{sections/02_method}
\input{sections/03_SimulationPlatform}
\input{sections/04_ExperimentalSetup}
\input{sections/05_Result}
\input{sections/07-conclusion}

\bibliographystyle{IEEEtran}
\bibliography{sample}

\end{document}

%% file: sections/00-abstract.tex
\begin{abstract}
Vehicle teleoperation has numerous potential applications, including serving as a backup solution for autonomous vehicles, facilitating remote delivery services, and enabling hazardous remote operations. However, complex urban scenarios, limited situational awareness, and network delay increase the cognitive workload of human operators and degrade teleoperation performance. To address this, the successive reference pose tracking (SRPT) approach was introduced in earlier work, which transmits successive reference poses to the remote vehicle instead of steering commands. The operator generates reference poses online with the help of a joystick steering and an augmented display, potentially mitigating the detrimental effects of delays. However, it is not clear which minimal set of sensors is essential for the SRPT vehicle teleoperation control loop.

This paper tests the robustness of the SRPT approach in the presence of state estimation inaccuracies, environmental disturbances, and measurement noises. The simulation environment, implemented in Simulink, features a 14-dof vehicle model and incorporates difficult maneuvers such as tight corners, double-lane changes, and slalom. Environmental disturbances include low adhesion track regions and strong cross-wind gusts. The results demonstrate that the SRPT approach, using either estimated or actual states, performs similarly under various worst-case scenarios, even without a position sensor requirement. Additionally, the designed state estimator ensures sufficient performance with just an inertial measurement unit, wheel speed encoder, and steer encoder, constituting a minimal set of essential sensors for the SRPT vehicle teleoperation control loop.

%The obtained results show that SRPT using estimated states performs similar to SRPT using actual states, even under various worst-case scenarios.

%\textcolor{blue}{Even with no GPS correction, SRPT performs well as EKF pose estimation is negligible in a small time-window.}
\end{abstract}


%% file: sections/01-introduction.tex
\section{Introduction}
\begin{figure}[ht]
\centering
    \includegraphics[width=1.0\columnwidth]{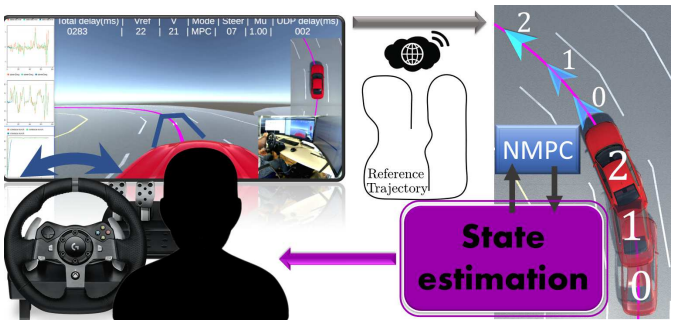}
    \caption{Graphical depiction of the SRPT approach utilized for direct vehicle teleoperation. The remote vehicle is provided with successive reference poses as it progresses forward. The control-loop includes a state-estimation block.}
    \label{fig:00_abstractImg}%
\end{figure}

\begin{figure*}[b!]
\centering
\subfloat[]{\includegraphics[width=0.8\textwidth]{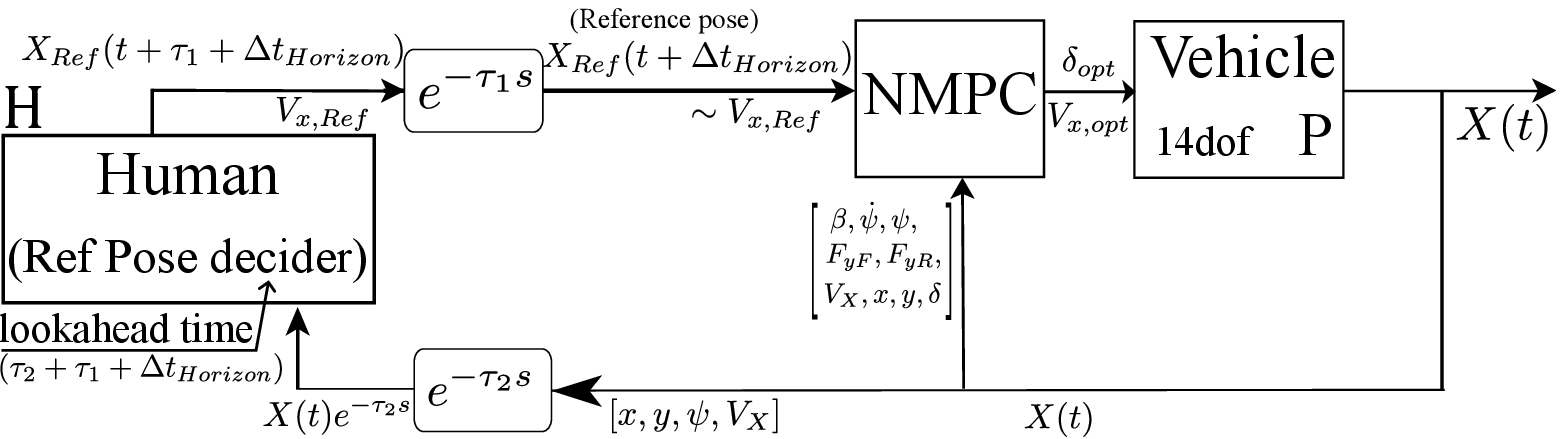}%
\label{fig:01_mpcScheme}}
\vfil
\subfloat[]{\includegraphics[width=0.8\textwidth]{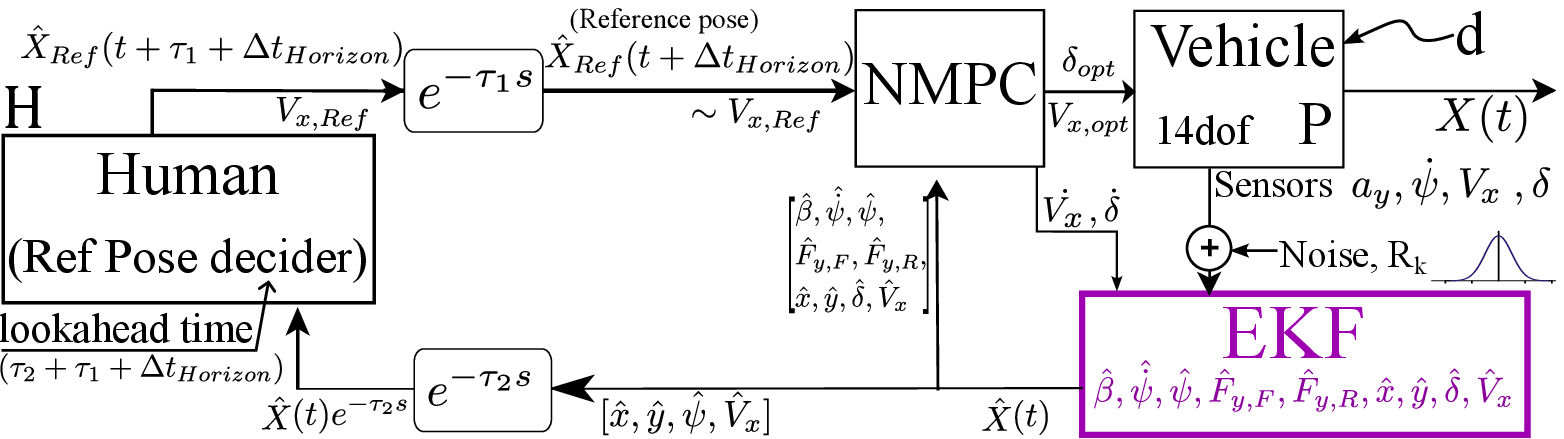}%
\label{fig:01_mpcSchemeEkf}}
\caption{SRPT block diagrams. (a) Without state estimation, its efficacy is discussed in \cite{jai2022_3}. (b) With state estimation, robustness assessment in presence of environmental disturbances and measurement noises is the prime focus of this work.}
\label{SRPTblocks}
\end{figure*}

\IEEEPARstart{T}{eleoperation} is the remote control of a device or a vehicle from a distance. This can be done using either wired communication or wireless communication. Here the vehicle is a mobile robot that can be controlled remotely, typically wirelessly.

Automated vehicles (AVs) are a promising technology for mobility in the future. Since their development has been a focus of academic and industrial study for many years, their introduction to the market is anticipated. Driverless cars with safety drivers can already be seen on public roads in some parts of the world. The utilization potential of AVs is, however, seriously threatened by lingering difficulties, required sophistication, or even as of yet undiscovered edge cases. A teleoperation system can apparently be considered a backup plan for AVs.
%By switching to remote operation, the AV can be supported by a human operator who is not in the vehicle but in a control-station.
The goal of teleoperation technology is to offer a secure and effective method to get over these restrictions anytime an autonomous driving (AD) function hits the limits of its Operational Design Domain (ODD) \cite{Domagoj2022}.
%The AV can resume its voyage in full automation after it has been returned to its nominal ODD \cite{Domagoj2022}.

\iffalse
Other applications of vehicle teleoperation can be:
\begin{itemize}
    \item Autonomous taxi service: Vehicles that can be summoned via a smartphone app and remotely driven to the passenger's location.
    \item Remote delivery service: Teleoperated vehicles for delivering goods, such as groceries or packages.
    \item Industrial equipment teleoperation: To remotely operate heavy machinery or other industrial equipment, which could be useful in hazardous environments or for tasks that require specialized skills.
    \item Disaster response: To remotely assess the damage, deliver supplies, and perform other necessary tasks.
    \item Military operations: Operations such as reconnaissance, improvised explosive device (IED) disposal, transporting equipment, medical evacuations, etc.
\end{itemize}
\fi

Ongoing shortcomings in vehicle teleoperation are:
\begin{enumerate}
  \item[!] Human-machine interaction: User experience of teleoperation systems, particularly in the areas of input modality (e.g. touch, gesture, voice), feedback (e.g. haptic, visual, auditory), and cognitive load \cite{4343985}.
  \item[!] Network latency: It introduces variable-delays and degradation in the performance of teleoperation, particularly in low-bandwidth or high-latency environments.
  \item[!] Safety and reliability: Robust fault detection, recovery mechanisms, and techniques for handling communication failures or malfunctions.
\end{enumerate}

While these issues continue to pose significant challenges, the objective of the approach discussed in this paper is to mitigate the impact of network latency in vehicle teleoperation. By addressing this specific challenge, we aim to improve the safety and effectiveness of teleoperation systems and move closer towards achieving reliable and efficient teleoperation of vehicles.

 Time delay reduces the speed and accuracy with which human operators can accomplish a teleoperation task \cite{7139813, Luck2006}. When delays are considerable, human operators tend to overcorrect steer, causing oscillations that degrade teleoperation performance and possibly destabilize the control-loop \cite{Sheridan1993, Gorsich2018EvaluatingMP}. Over the years, different teleoperation concepts for road vehicles have been developed and researched. Domagoj Majstorovic et al. \cite{Domagoj2022} suggest a taxonomy for vehicle teleoperation approaches:

\begin{enumerate}
  \item[$\blacksquare$] Direct control: The human operator sees the sensor data and transmits control signals, such as steer, throttle, etc. It suffers from reduced situational awareness and transmission latency \cite{Georg2020, Mutzenich2021, Hoffmann2021, chucholowski2016, Tang2014, Argelaguet2020, Georg2019, Georg2020_2, jai2022}.
  \item[$\blacksquare$] Shared control: A shared controller inside the vehicle assesses the operator commands as per current surrounding and manipulates them to avoid an imminent collision \cite{Saparia2021, Anderson2013, Schimpe2020, Qiao2021, Schitz2021, Justin2017}. It improves safety but still suffers from latency.
  \item[$\blacksquare$] Trajectory or waypoints guidance: The vehicle tracks the path and speed profile generated by the operator. Network latency doesn't impact the performance, but real-time generation of path and speed profile is difficult \cite{Gnatzig2012, Hoffmann2022, Kay1995, Bjornberg1504690, Schitz2021_2}.
  \item[$\blacksquare$] Interactive path planning: The perception module of vehicle computes optimal paths and the operator confirms one of them to follow. It circumvents network latency but requires a functional set of AD modules such as perception, trajectory planning, etc. \cite{Hosseini2014, Schitz2021_3}
  \item[$\blacksquare$] Perception modification: The human operator supports the AD perception module by identifying false-positive obstacles. It majorly depends upon the AD modules \cite{Feiler2021}.
\end{enumerate}

\subsection{Previous Work}
Previously, we proposed \cite{jai2022_2} and evaluated \cite{jai2022_3} the novel SRPT approach for vehicle teleoperation. It combines the benefits of direct control and waypoint guidance concepts. The human operator online generates waypoints using joystick steering by manipulating an augmented lookahead point on the display. Eventually, the vehicle controller drives the vehicle. In this earlier work, a simulation environment is used, and so the vehicle controller is aware of the (perfect) actual vehicle states. But in the real world, all the states are not readily available due to obvious measurement limitations. Robustness of the novel SRPT approach for vehicle teleoperation has to be assessed in presence of state estimation accuracies. Furthermore, a minimal but sufficient set of sensors has to be identified for the state estimation requirement of SRPT.

\subsection{Related work - Vehicle state estimation}
Bersani Mattia et al. \cite{BERSANI2021103662} formulated an unscented Kalman filter (UKF) with kinematic single-track vehicle model to estimate vehicle pose and velocities. Its plant model input are IMU measurements and not the steering angle. The measurement model consists of GPS position, GPS speed and speed measured by wheel encoders. It doesn't estimate tire shear forces.
Doumiati Moustapha et al. \cite{Doumiati2011} formulated an extended Kalman filter (EKF) with a four-wheel vehicle model to estimate the side-slip angle and lateral tire forces. Its plant model input is the steering angle. Its measurement model consists of yaw rate, vehicle velocity, longitudinal, and lateral accelerations. In our paper, we use the single-track version of this EKF formulation.

\subsection{Contribution of Paper}

\begin{boxlabel}
\item Performance assessment of the SRPT vehicle teleoperation approach in the presence of state estimation inaccuracies.
\item Identification of the minimum set of sensors required for state estimation in the SRPT vehicle teleoperation control loop, even under challenging scenarios involving difficult maneuvers, adverse environmental conditions, and various types of measurement noises.
\item Highlights that this approach does not require a GPS sensor, despite relying on sending reference poses to the remote vehicle.
\end{boxlabel}

\iffalse
\vspace{-3pt}
\begin{figure}[H]
%    \centering
    \includegraphics[width=0.6\textwidth]{figures/00_AbstractImg.eps}
    \caption{Integration of Simulink and Unity for human-in-loop vehicle teleoperation experiments.}
    \label{fig:x 00_AbstractImg}
\end{figure}
\fi

\subsection{Outline of Paper}
The paper is organized as follows. Section \ref{sec2} presents the skeleton of the EKF state-estimator. Section \ref{sec2D} reiterates the role of human operator in SRPT approach. Section \ref{sec3} provides an overview of the simulation platform. Section \ref{sec4} discusses the experimental structure. Section \ref{sec5} presents and discusses the results. Section \ref{sec6} concludes with the work summary.

%% file: sections/02_method.tex
\section{Method}\label{sec2}
\noindent The SRPT approach proposed in \cite{jai2022_3}, requires vehicle states. The block diagram of SRPT approach is shown in figure \ref{fig:01_mpcScheme}. Considering the single-track vehicle model for state estimation, the states are side-slip angle ($\beta$), vehicle yaw-rate ($\dot{\psi}$), vehicle heading ($\psi$), lateral force at front axle ($F_{y, F}$), lateral force at rear axle ($F_{y, R}$), vehicle longitudinal speed ($V_x$), position of vehicle CG ($[x; y]$), and equivalent steer angle at front axle ($\delta$):
\begin{equation}
X=\left[\:\beta \:, \:\dot{\psi} \:, \:\psi \:, \:F_{y, F} \:, \:F_{y, R} \:, \:V_x, \:x \:, \:y \:, \:\delta \:\right]^T\label{eq states} \;\;\;\;.
\end{equation}
The NMPC block optimizes for vehicle steer-speed commands, to keep the vehicle motion in sync with the received successive reference poses \cite{jai2022_3}. It also accounts for input constraints such as steer-rate constraint (actuator limitation) and acceleration constraint (for passenger comfort):
\begin{align}
\left[\dot{\delta}_{min}=-20^{\circ}/s\right] &\leq \dot{\delta} \leq \left[\dot{\delta}_{max}=+20^{\circ}/s\right] \label{eq inputConstraints1}\\
\:\:\:\:\left[a_{min}=-3m/s^2\right] &\leq a \leq \left[a_{max}=1m/s^2\right] \label{eq inputConstraints2}
\end{align}
Its outputs are, the steer angular velocity ($\dot{\delta}$), and vehicle acceleration ($a$).
%Consequently, optimized steer angle ($\delta_{opt}$) and vehicle speed ($V_{x, opt}$) can be commanded to the low-level vehicle controller.

Figure \ref{fig:01_mpcSchemeEkf} shows the incorporation of a state-estimator in the form of EKF. Its prediction model consists of the single-track version of the four-wheel vehicle model given in \cite{Doumiati2011} and it is given by

\begin{equation}
\dot{X}=\left[\begin{array}{c}
\frac{1}{m V_x}\left(F_{y, F} \cos \delta+F_{x, F} \sin \delta+F_{y, R}\right)-\frac{\beta \cdot a}{V_x}-\dot{\psi} \\
\frac{1}{I_Z}\left[\left(F_{y, F} \cos \delta+F_{x, F} \sin \delta\right)l_F - F_{y, R}\:l_R\right] \\
\dot{\psi} \\
%\frac{V_x}{\lambda}\left[\zeta_F\,C_{\sigma, F} \:\sigma_{F}-F_{y, F}\right] \\
%\frac{V_x}{\lambda}\left[\zeta_R\,C_{\sigma, R} \:\sigma_{R}-F_{y, R}\right] \\
\frac{V_x}{\lambda}\left[C_{\sigma, F} \:\sigma_{F}-F_{y, F}\right] \\
\frac{V_x}{\lambda}\left[C_{\sigma, R} \:\sigma_{R}-F_{y, R}\right] \\
\dot{V_x} \\
V_x \left(\cos \psi - \sin \psi \; \tan \beta \right) \\ 
V_x \left(\sin \psi + \cos \psi \; \tan \beta \right) \\ 
\dot{\delta}
\end{array}\right].\label{eq statesDot}
\end{equation}
%\vspace{-6pt}

$[C_{\sigma, F}, C_{\sigma, R}]$ are the lumped cornering stiffness of front and rear axles respectively. [$m_f, m_R$] are distribution of vehicle mass on front and rear axle based on [$l_F, l_R$] respective distances of axles from CG. To make it applicable for zero vehicle speed, wherever the $V_x$ is in the denominator in (\ref{eq statesDot}), it is substituted by $max(0.01, V_x)$.

For simplicity, relaxation length phenomenon, rolling resistance and aerodynamic drag are ignored in longitudinal force development and thus given by
\begin{equation}
\begin{aligned}
&F_{x, F}= 
\begin{cases}
    \:\:\:\:\:\:\:\:m\,a,& \text{if } a\geq 0\\
    \:\:\:\:\gamma\,m\,a               & \text{otherwise},
\end{cases}
%\\
%&F_{x, R}= 
%\begin{cases}
%    \:\:\:\:\:\:\:\:\:\: 0,& \text{if } a\geq 0\\
%    (1-\gamma)\,m\,a              & \text{otherwise},
%\end{cases}
\end{aligned}\label{eq Fxf}
\end{equation}
where $\gamma$ is the braking bias.

%$[\zeta_F, \zeta_R]$ are the reduction factor for cornering stiffness in presence of longitudinal forces and are given by
%\begin{equation}
%\begin{array}{l}
%\zeta_F=\sqrt{1-\left(\frac{F_{x, F}}{ \:m_{F}\:g}\right)^{2}}\\
%\zeta_R=\sqrt{1-\left(\frac{F_{x, R}}{ \:m_{R}\:g}\right)^{2}}
%\end{array}
%\end{equation}\label{eq zeta}

$[\sigma_{F}, \sigma_{R}]$ are the tires slips given by
\begin{equation}
\begin{array}{l}
\sigma_{F}\simeq \tan \delta-\beta-\dot{\psi} \frac{l_F}{V_x} \\
\sigma_{R}\simeq \:\:\:\:\:\:\:\:\:\:\:-\beta+\dot{\psi}\frac{ l_R}{V_x}
\end{array}
\end{equation}\label{eq slips}

\begin{figure}[h]
\centering
    \includegraphics[width=0.6\columnwidth]{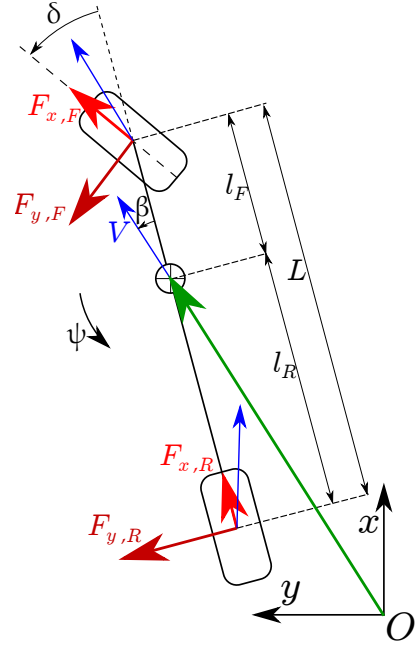}
    \caption{Single-track vehicle model. Reprinted with permission from Ref. \cite{jai2022_2}}
    \label{fig:02_singleTrack}%
\end{figure}

Parameters used in the state estimation correspond to a FWD typical passenger vehicle and mentioned in Table \ref{tab:vehicleParameters}.
\begin{table}[ht]
\centering
\caption{Vehicle parameters for the single-track model.}
\label{tab:vehicleParameters}
\begin{tabular}{cc}
\hline
\textbf{Parameter}                & \textbf{Value}             \\ \hline
$m$                               & 1681\:kg                   \\ \hline
$I_z$                             & 2600\:kg\,s$^2$            \\ \hline
$[m_F;\:m_R]$                     & [871.6;\:809.4]\:kg        \\ \hline
$[l_F;\:l_R]$                     & [1.3;\:1.4]\:m             \\ \hline
$[C_{\sigma, F};\:C_{\sigma, R}]$ & [1.057e+05;\:1.050e+05]\:N \\ \hline
$\lambda$ (Relaxation\,length)    & 0.3\:m                     \\ \hline
$\gamma$ (Braking\,bias)          & 0.6                        \\ \hline
\end{tabular}
\end{table}

\iffalse
\subsection{Prediction model}
\noindent Euler-method is used to discretize the dynamics given in (\ref{eq statesDot}), the prediction step is given by

\begin{align}
\hat{X}_{k|k-1}&= f_{k-1} (\hat{X}_{k-1|k-1, \;u_k}) = \dot{X}_{k-1|k-1, \;u_k}dt + \hat{X}_{k-1|k-1} \label{eq predictStateEstimate}\\
P_{k \mid k-1} &=F_k P_{k-1 \mid k-1} F_k^T+Q_k \label{eq predictCovarianceEstimate}
\end{align}

Here $\hat{X}_{k|k-1}$ is predicted state estimate, $[dt=0.001s]$ is the prediction time step, $Q_k$ is covariance realted to each prediction step (its tuning is discussed in section~\ref{sec:predictionNoise}), and $P_{k \mid k-1}$ is predicted covariance estimate covariance. The state transition matrix ($F_k$) is given by
\begin{equation}
F_k=\left.\frac{\partial f}{\partial X}\right|_{\hat{X}_{k-1 \mid k-1}, u_k},
\end{equation}
considering the definition of $f()$ given in (\ref{eq predictStateEstimate}), $F_k$ can be calculated as
\begin{equation}
F_k=\left.\frac{\partial \dot{X}}{\partial X}\right|_{\hat{X}_{k-1 \mid k-1}, u_k} \cdot dt + I\;\;\;\;.
\end{equation}
\fi

Discretization steps are the usual first-order approximation with a time step of $0.001s$, as also presented in the appendix of \cite{Cheng2019}. Tuning of prediction noise covariance, $Q_k \in \mathbb{R}^{9\times9}$, is discussed in the section below.
 %~\ref{sec:predictionNoise}.
%It has to be noted down that the in state transition matrix formulation, $F_{x, F:R}$ and $\gamma_{F:R}$ are independent of states.

\begin{figure}[ht]
%\centering
    \includegraphics[width=1\columnwidth]{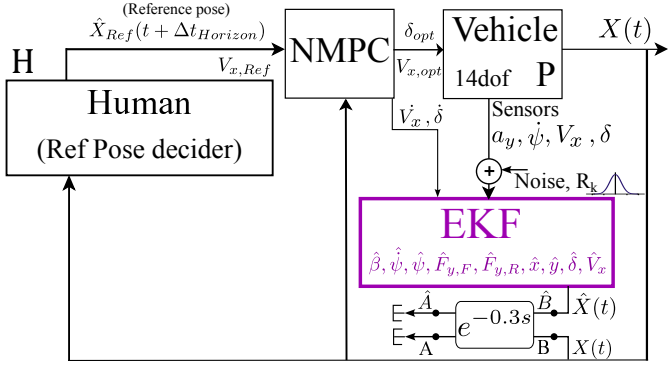}
    \caption{Block diagram used to optimally tune the prediction covariance $Q_k$, by minimizing prediction error of relative pose at $\hat{B}$ wrt pose at $\hat{A}$. The trajectory is traversed at $V_{Ref}=22 km/h$.}
    \label{fig:03_tuningQ}%
\end{figure}

\subsection{Prediction noise, $Q_k$} \label{sec:predictionNoise}
$Q_k$ is assumed to be a diagonal matrix with respective variances at the diagonal. To tune $Q_k$, the vehicle performs one lap over the whole trajectory given in figure \ref{fig:x 07_Trajectory}. Actual vehicle states and (zero mean Gaussian) noisy measurements are recorded. Referring to figure \ref{fig:03_tuningQ}, to obtain relative poses for a time window of $0.3s$ (max round-trip delay), both estimated pose and actual pose are delayed by $0.3s$. Comparison is performed between estimated relative poses and actual relative poses. $Q_k$ is optimally tuned offline, aiming minimum prediction error in relative poses, the cost function is given by (\ref{eq costFmincon}). The aim is to tune the variances corresponding to just $6$ out of $9$ states, i.e., $\left[\:\beta \:, \:\dot{\psi} \:, \:F_{y, F} \:, \:F_{y, R} \:, \:\delta \:, \:V_x\right]$. This is because other states are unobservable with the measurement model given in (\ref{eq measeurementModel}). 
The nonlinear multivariable solver of Matlab, \textit{fmincon}, is used to optimize iteratively the variances corresponding to these $6$ states.
\begin{multline}
J=
%W_{Position} RMS\left(\left\|{}^B\hat{P}_A - {}^BP_A \right\|_2 \right)+ 
W_{Position} RMS\left(\left\|\vec{\hat{B}}_{\hat{A}} - \vec{B}_{A} \right\|_2 \right)+ \\
%W_{Heading}  RMS\left(\left\|{}^B\hat{\psi}_A - {}^B\psi_A \right\|_2 \right)
W_{Heading}  RMS\left(\left\|\psi_{\hat{A}}^{\hat{B}} - \psi_A^B \right\|_2 \right)
\label{eq costFmincon} \;\;\;\; .
\end{multline}
$\vec{\hat{B}}_{\hat{A}}$ - Predicted relative position of $\hat{B}$ in $\hat{A}$ frame (a vector of 2 elements $[\hat{x};\hat{y}]$)
\\
$\vec{B}_{A}$ - \,\,\,\,\,\,\,Actual relative position of ${B}$ in ${A}$ frame (a vector of 2 elements $[x;y]$)
\\
\vspace{1mm}
$\psi_{\hat{A}}^{\hat{B}}$ - Predicted relative heading of $\hat{B}$ in $\hat{A}$ frame (a scalar)
\\
\vspace{1mm}
$\psi_A^B$ - \,\,\,\,\,\,\,Actual relative heading of ${B}$ in ${A}$ frame (a scalar)
\\
$W_{Position}=1$ \,\,\,\,\,\,\,\,\,\,\,\,\,\,\,\,\,\,\, - Weight for the position error
\\
$W_{Heading}=1e^{-2}$ \,\,\,\,\,\,\,\,\,\,\,\,- Weight for the heading error
\\
$RMS()$ is performed over the whole trajectory. After the solver optimization, the tuned process covariance ($Q_k$) is given by

\begin{equation}
\begin{split}
Q_k = \frac{\left(
    diag\left[
    \begin{aligned}
         & \:0.0125 \:, \:0.0011 \:,\:1 \:,\:0.3162 \:,\\
         & \:0.3415 \:, \:0.0008 \:,\:1 \:,\:1 \:,\:2e-5 \:\\
    \end{aligned}
    \right]
    \right)^2}{10}
\end{split}
\label{eq processCovariance}
\end{equation}

E.g. process noise for state $\beta$ is $\mathcal{N}\left(0,\,0.0125^{2}/10\right)$ and so on for other states. It is presented as a factor of $10$, so that $std$ of the vehicle states which are as it is in measurement vector, can be readily compared, as there are $10$ prediction steps in between two measurements. Process noise corresponding to unobservable states, $[x, y, \psi]$, are arbitrarily assigned $\mathcal{N}\left(0,\,1^{2}/10\right)$, as estimation of their evolution is unimportant.

\subsection{Measurement model} \label{sec:measurementNoise}
\noindent A minimal set of sensors, comprising a virtual IMU, a speed encoder and a steer encoder is considered. Intending to measure only lateral acceleration at CG ($a_{y, meas}$), angular velocity at CG ($\dot{\psi}_{meas}$), longitudinal speed ($V_{x, meas}$) and steer angular position ($\delta_{meas}$). Hence the measurement vector is given as

\begin{equation}
Z_{k}=\left[\:a_{y, meas} \:, \:\dot{\psi}_{meas} \:, \:V_{x, meas} \:, \:\delta_{meas}\right]^T\label{eq measeurementVector} \;\;\;\; ,
\end{equation}

and the measurement model is given as

\begin{equation}
h\left(X\right)=\left[\begin{array}{c}
\frac{1}{m}\left(F_{y, F} \cos \delta+F_{y, R}\right) \\
\dot{\psi} \\
V_{x} \\
\delta
\end{array}\right] \;\;\;\; . \label{eq measeurementModel}
\end{equation}

\iffalse
\begin{equation}
h\left(X_k\right)=\left[\begin{array}{c}
\frac{1}{m}\left(F_{y, F, k} \cos \delta+F_{y, R,k}\right) \\
\dot{\psi}_k \\
V_{x, k} \\
\delta_k
\end{array}\right] \;\;\;\; . \label{eq measeurementModel}
\end{equation}

The observation matrix ($H_k$) is given by
\begin{equation}
H_k=\left.\frac{\partial h}{\partial X}\right|_{\hat{X}_{k \mid k-1}} \;\;\;.
\end{equation}

The update steps are given as
\begin{align}
& V_k=Z_k-h\left(\hat{X}_{k \mid k-1}\right) \\
& S_k=H_k P_{k \mid k-1} H_k^T+R_k \\
& K_k=P_{k \mid k-1} H_k^T S_k^{-1} \\
& \hat{X}_{k \mid k}=\hat{X}_{k \mid k-1}+K_k V_k \\
& P_{k \mid k}=\left(I-K_k H_k\right) P_{k \mid k-1} \;\;\;\; .
\end{align}

Here $V_k$ is innovation, $S_k$ is innovation covariance, $R_k$ is measurement covariance (further discussed in section~\ref{sec:noises}), $K_k$ is Kalman gain, $\hat{X}_{k \mid k}$ is updated state estimate, and $P_{k \mid k}$ is updated covariance estimate.
\fi

The update steps are incapable to correct the vehicle pose ($x, y, \psi$) because these states are unobservable with the measurement model given in (\ref{eq measeurementModel}). Consequently, these states have the tendency to diverge with time. Besides of this divergence, robustness of SRPT approach majorly depends upon the accuracy of (relative) pose estimation for a time window equivalent to the round trip network delay, further discussed in section~\ref{sec5A}.

%\subsection{Measurement noise} \label{sec:noises}
Sampling time-step is $0.010 s$. Taking advantage of using a simulation framework, zero-mean Gaussian noise is added to the measurements. Measurement covariance respective to measurement vector (\ref{eq measeurementVector}) is given by
\begin{equation}
R_k=diag\left(\left[\:0.112^2 \:, 0.005^2 \:, \:0.083^2 \:, \:0.003^2\right]\right) \label{eq measeurementCovariance} \;\;\;\; .
\end{equation}

% Add this in a note instead
Standard deviations corresponding to $a_{y, meas}$ and $\dot{\psi}_{meas}$ are taken from a static experiments performed with Bosch 5-axis IMU. Standard deviations corresponding to $V_{x, meas}$ and $\delta_{meas}$ are taken from speed and steer encoders.

\subsection{Human model}\label{sec2D}
\noindent In contrast to conventional vehicle teleoperation, where human operator transmits steer-throttle commands to the remote vehicle, in SRPT approach, human operator transmits reference poses to be followed by the remote vehicle. These reference poses are generated and transmitted in such a way that, when the vehicle receives them, they resemble those ahead of it. Look-ahead time of $[1+\tau_1+\tau_2]s$ is considered for reference pose generation. As a result, the vehicle receives reference-poses that are approximately $1s$ ahead of it. The $1s$ horizon is chosen arbitrary, taking into account that, 
in general, a driver steers a vehicle based on an upcoming vehicle position. Eventually, the same time horizon of $\Delta t_{Horizon}=1s$ is considered for the NMPC block to optimize for vehicle steer-speed commands. 

\begin{figure}[h]
%\centering
    \includegraphics[width=1\columnwidth]{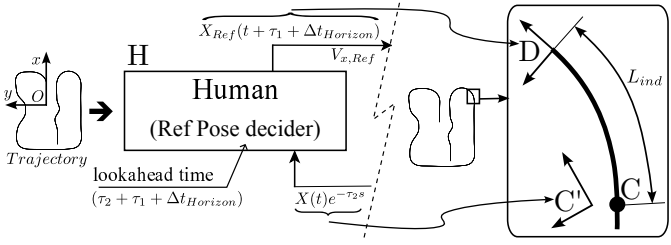}
    \caption{Working principle of the human model block for the case of perfectly known states (Figure \ref{fig:01_mpcScheme}). Its task is to choose the future reference pose based on the received vehicle pose and look-ahead distance.}
    \label{fig:04_humanModelPerfectStates}%
\end{figure}

The task of the human model block is to transmit information that informs the vehicle about its aiming direction. Referring to figure \ref{fig:04_humanModelPerfectStates}, human model block receives delayed vehicle states, $X(t)e^{-\tau_2 s}$, which consists of vehicle pose, ${P}^{C'}_O$. It is the delayed vehicle pose in global reference frame, $O$. Being aware of the whole trajectory, the human model block first finds the closest point $C$ on the reference trajectory. Then it finds the point $D$, which is $L_{ind}$ distance ahead of point $C$. The $L_{ind}$ is the look-ahead distance govern by below relation:
\begin{equation}
L_{ind} = V_x\cdot\tau + \max(V_x\cdot \Delta t_{Horizon}, \;\;l_F) \label{eq lookaheaDistance}\;\;\;\; .
\end{equation} 

It is lower bounded by $l_F$, the front axle distance from CG. It is linearly proportional to the round trip delay ($\tau$) and to the vehicle speed ($V_x$). The first term tries to compensate for round-trip delay, and the second term aims to generate the terminal condition for the NMPC horizon.

\begin{figure*}[!b]
\centering
    \includegraphics[width=0.8\textwidth]{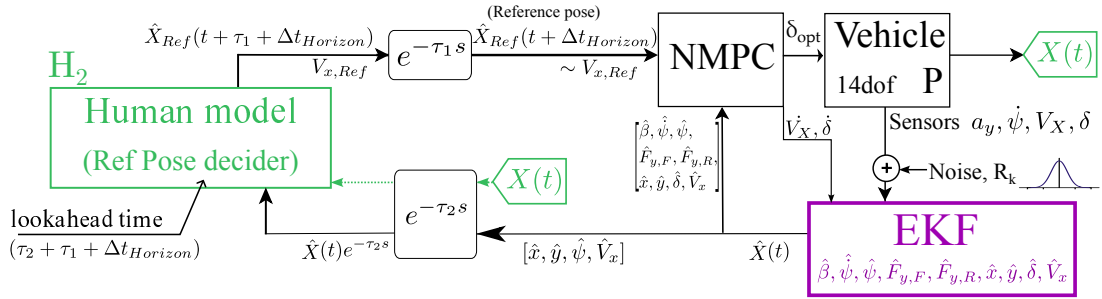}
    \caption{SRPT block diagram with state estimator and human model block. To compensate for the absence of visual feedback to human model block, it has additional feedback (highlighted in green) of delayed but actual vehicle state, $X(t)$.}
    \label{fig:05_humanModelWithEKF}%
\end{figure*}

%\color{blue}{
There are two ways in which the aiming direction can be transmitted to the vehicle:
\begin{enumerate}
%\color{blue}
\item Transmit the relative reference pose, ${P}^{D}_{C'}$. It is relative position and heading of pose-D with respect to pose-C$'$, it acts as a correction term, which tries to bring the vehicle close to the desired trajectory. Upon receive of this relative-pose, the vehicle first estimates how much it has already traveled during the round-trip delay and how much more it has to travel. This estimation is possible, as messages are timestamped.

\item Transmit the global reference pose. Transmit the reference pose, $X_{Ref}$, in global reference frame.
\begin{equation}
X_{Ref}={P}^{C'}_O + {P}^{D}_{C'} = {P}^{D}_{O} \label{eq humanPerfectStates}\;\;\;\; .
\end{equation} 

\end{enumerate}
%\textcolor{blue}{
For this paper, the second way is adopted. It eliminates the vehicle task of estimating how much it has traveled during the round-trip delay.
%Since the focus of this paper is the effect of vehicle state estimation error, which is accounted in the simulation environment, this approach is reasonable. 
%}
%The reference pose to be transmitted by the human model is given by:
%As it can be noticed that the point $C'$ is offset from the desired trajectory, ${P}^{D}_{C'}$ is acting as a correction term, which tries to bring the vehicle close to the desired trajectory.

In case when the human block receives predicted states $\left( \hat{X}\left(t \right) e^{-\tau_2 s} \right)$, two further cases has to be discussed.

\begin{enumerate}[label=(\roman*)]
\item Human in the loop - The human block is a human operator. The operator receives the delayed predicted vehicle state as well as he/she sees the delayed visual environment. With these modality the correction term, ${P}^{D}_{C'}$, can be decided by the operator. In our previous work \cite{jai2022_3}, this correction term is getting generated online using steering joystick, briefly represented by below relation

\begin{equation}
\hat{X}_{Ref}=\hat{P}^{C'}_O + \Delta P_{Joystick} \label{eq humanPredictedStates}\;\;\;\; .
\end{equation} 

$\hat{P}^{C'}_O$ - It denotes the predicted vehicle pose, readily available in $\hat{X}\left(t \right) e^{-\tau_2 s} $
\\
$\Delta P_{Joystick}$ - It is the correction term generated by the augmented steer indicator marker on the visual interface with help of joystick steering \cite{jai2022_3}.
\\

\item Human model in the loop - Human is substituted by a human model that only receives delayed predicted vehicle states. To compensate for the absence of visual environment, this block has to be additionally fed with delayed but actual vehicle states, as shown in green in figure \ref{fig:05_humanModelWithEKF}. This is because the predicted vehicle poses diverge from the actual vehicle poses (due to measurement model limitations discussed section
~\ref{sec:measurementNoise}
).
Diverging vehicle states can't be compared with the desired trajectory. To get the closest point $C$, actual vehicle states are required. %That's why the human model block is additionally fed with delayed but actual vehicle states.
Passing actual states makes the human model block perfect, which supports the main focus of this paper, which is, the performance assessment of SRPT approach in vehicle teleoperation in presence of state estimation error. The teleoperation loop contains only state prediction error, no human error.

The reference pose to be transmitted by the human model block is given by:

\begin{equation}
\hat{X}_{Ref}=\hat{P}^{C'}_O + {P}^{D}_{C'} \label{eq humanModelPredictedStates}\;\;\;\; .
\end{equation} 

${P}^{D}_{C'}$ - It is the relative pose of $D$ on the trajectory with respect to the delayed actual vehicle pose ($C'$). The correction term is same as in (\ref{eq humanPerfectStates}). This is available, as the human model block is additionally fed with delayed actual vehicle states.
%\subsection{NMPC block}\label{nmpc}
%\noindent Vehicle receives the reference pose $\hat{X}_{Ref}$, due to variable network delay it may not be exactly $1s$ ahead of vehicle but it is around $1s$ ahead of vehicle. %To align $\hat{X}_{Ref}$ with vehicle path, NMPC block explained in \cite{jai2022_3}, optimizes for steer and speed commands.
\end{enumerate}

%% file: sections/03_SimulationPlatform.tex
\section{Simulation Platform}\label{sec3}

\begin{figure*}[!t]
    \centering
    \includegraphics[width=0.8\textwidth]{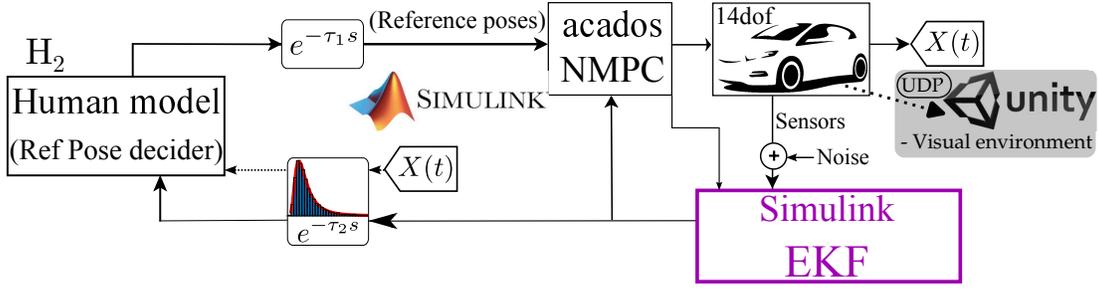}
    \caption{Simplified block diagram of the simulation platforms set up on Simulink. Unity has no role in simulation, it is just to display the manoeuvres.}
    \label{fig:x 06_simulationPlatform}
\end{figure*}

\begin{table*}[!t]
\centering
\caption{Description of the blocks used in the simulation platform.}
\begin{tabular}{|l|l|l|}
\hline
\multicolumn{1}{|c|}{\textbf{Block}} & \multicolumn{1}{c|}{\textbf{Description}}                     & \multicolumn{1}{c|}{\textbf{Rate}} \\ \hline
Vehicle 14dof                        & A 14dof vehicle model to simulate a real vehicle              & 1000 Hz                            \\ \hline
Sensors                              & It is a channel extracting IMU and encoders observations      & 100 Hz                             \\ \hline
EKF                                  & It is a Simulink Extented Kalman filter block                 & 1000/100 Hz                        \\ \hline
$e^{-\tau_2 s}$                      & Variable network downlink delay                               & 30 Hz                              \\ \hline
Human model                          & Ref Pose decider logic                                        & 30 Hz                              \\ \hline
$e^{-\tau_1 s}$                      & Constant network uplink delay, $\tau_1 = 0.060s$              & 30 Hz                              \\ \hline
NMPC                                 & Non-linear model predictive controller, Acados toolkit        & 50 Hz                              \\ \hline
Unity                                & An external block, to visualize the real vehicle maneuvers & 100 Hz                             \\ \hline
\end{tabular}
\label{tab:blockDescriptions}
\end{table*}

\noindent A faster than real-time, vehicle teleoperation simulation test platform is developed using Simulink + Unity3D to emulate the network delayed vehicle teleoperation system, as shown in figure \ref{fig:x 06_simulationPlatform}. The Simulink output is connected to Unity only to provide visuals of the real vehicle manoeuvres. 

Further descriptions of the blocks are mentioned in table \ref{tab:blockDescriptions}. The table also mentions the working rate of each block, different working rate of each block aligns with reality, where each module has its own working capabilities. The EKF block has prediction step of $dt = 0.001s$ and measurement update at each $0.010s$. The blocks $e^{-\tau_2 s}$, Human model, and $e^{-\tau_1 s}$  work synchronously with each other at $30 Hz$, to simulate the usual discrete nature of video streaming to the control station. The downlink delay ($\tau_2$) is considered variable delay to simulate usual network delays but the uplink delay ($\tau_1$) is considered a constant of $0.060s$ due its lower magnitude and lower variability. For downlink delay, generalized extreme value distribution, $GEV(\xi=0.29, \mu=0.200, \sigma=0.009)$ is used~\cite{Zheng2020, jai2022}. Here, $\xi$ is the shape parameter, $\mu$ is the location parameter and $\sigma>0$ is the scale parameter. Positive $\xi$ means that the distribution has a lower bound $(\mu-\frac{\sigma}{\xi})\approx0.169$~s$\:(>0)$ and a continuous right tail based on the extreme value theory. 
This probability distribution keeps variable downlink delay in range of $0.169s-0.300 s$.

%% file: sections/04_ExperimentalSetup.tex
\begin{table*}[t]
\centering
\caption{Sources of measurement noise}
\begin{tabular}{|l|l|}
\hline
\multicolumn{1}{|c|}{\textbf{Noise sources}}                                           & \multicolumn{1}{c|}{\textbf{Description}}                                                                                                                                                                                                     \\ \hline
Gaussian noise                                                                         & AWGN; $R_k=diag\left(\left[\:0.112^2 \:, 0.005^2 \:, \:0.083^2 \:, \:0.003^2\right]\right)$                                                                                                                                                   \\ \hline
$gainV = 1.05$                                                                         & \begin{tabular}[c]{@{}l@{}}Vehicle speed = rolling speed $\times$ rolling radius.\\ Inaccuracy of rolling radius act as a gain form of disturbance.\end{tabular}                                                                              \\ \hline
$bias\delta = 0.5^{\circ}$                                                             & Offset mounting of steer position encoder.                                                                                                                                                                                                    \\ \hline
$3^{\circ}tiltedImu$                                                                   & \begin{tabular}[c]{@{}l@{}}$5^{\circ}$ tilted mounting of IMU in clockwise roll direction;\\ $a_{y, noisy} = a_ycos(3^{\circ}) - 9.8sin(3^{\circ})\;\;\;\;\;\;\; ;\;\;\;\;\;\;\;\;\dot{\psi}_{noisy} = \dot{\psi}cos(3^{\circ})$\end{tabular} \\ \hline
$6^{\circ}tiltedImu$                                                                   & \begin{tabular}[c]{@{}l@{}}$6^{\circ}$ tilted mounting of IMU in clockwise roll direction;\\ $a_{y, noisy} = a_ycos(6^{\circ}) - 9.8sin(6^{\circ})\;\;\;\;\;\;\; ;\;\;\;\;\;\;\;\;\dot{\psi}_{noisy} = \dot{\psi}cos(6^{\circ})$\end{tabular} \\ \hline
\begin{tabular}[c]{@{}l@{}}$C_{F, over} = 1.2C_F$\\ $C_{R, over}= 1.2C_R$\end{tabular} & Use of overly estimated cornering stiffness for state estimation.                                                                                                                                                                             \\ \hline
\end{tabular}
\label{tab:worstCaseScenarios}
\end{table*}

\begin{figure}[!h]
    \centering
    \includegraphics[width=0.84\columnwidth]{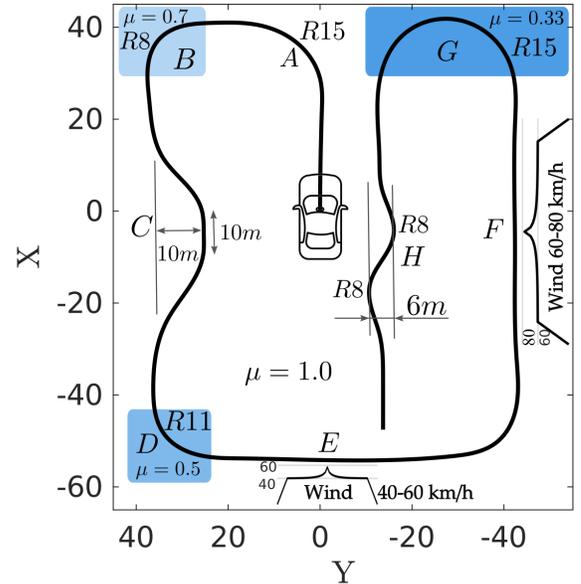}
    \caption{Track contains various sections A-H of difficult manoeuvres and worst-case environmental conditions.}
    \label{fig:x 07_Trajectory}
\end{figure}

\begin{figure*}[!t]
\centering
\subfloat[]{\includegraphics[width=0.43\textwidth]{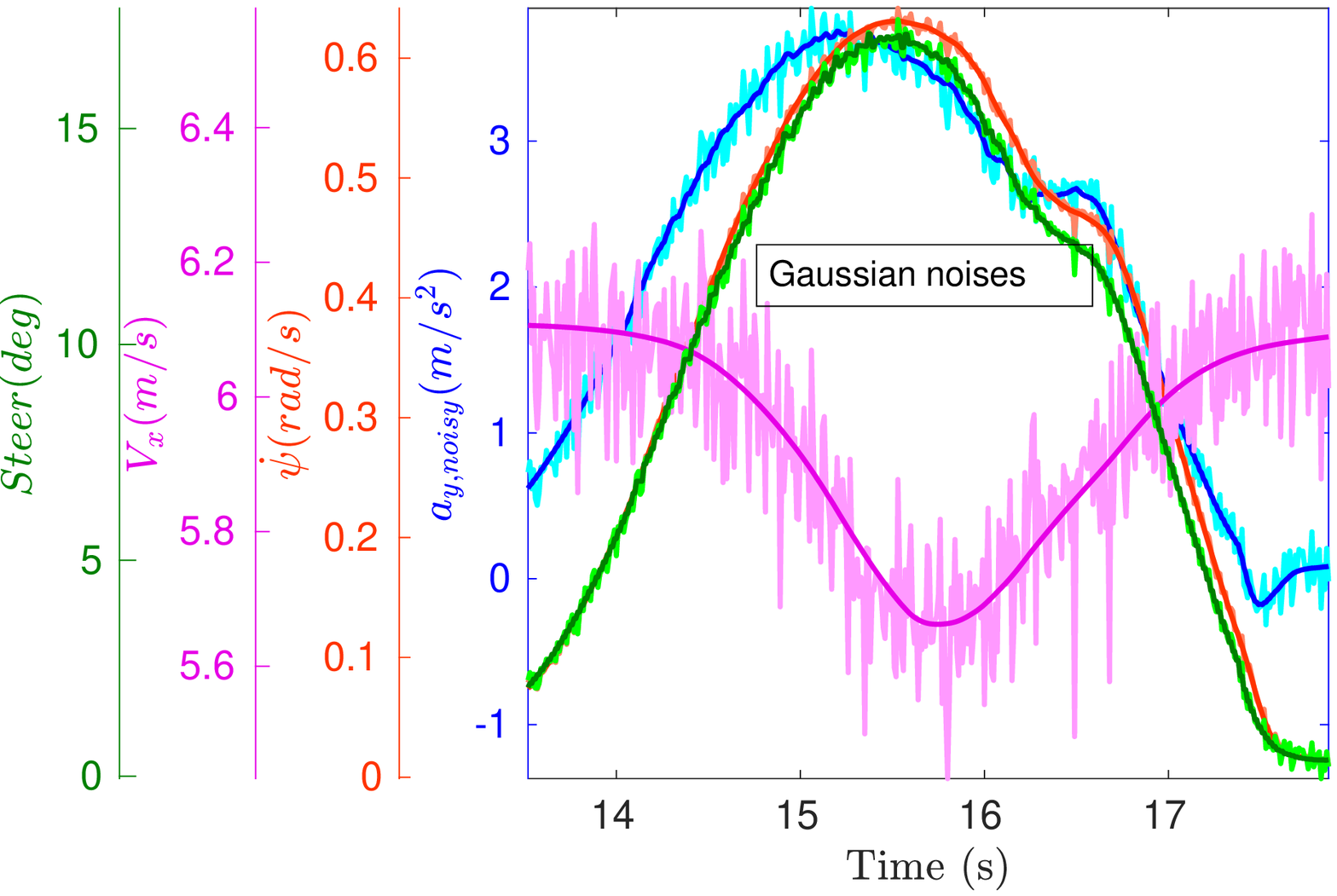}%
\label{fig:0804}}
\hfil
\subfloat[]{\includegraphics[width=0.274\textwidth]{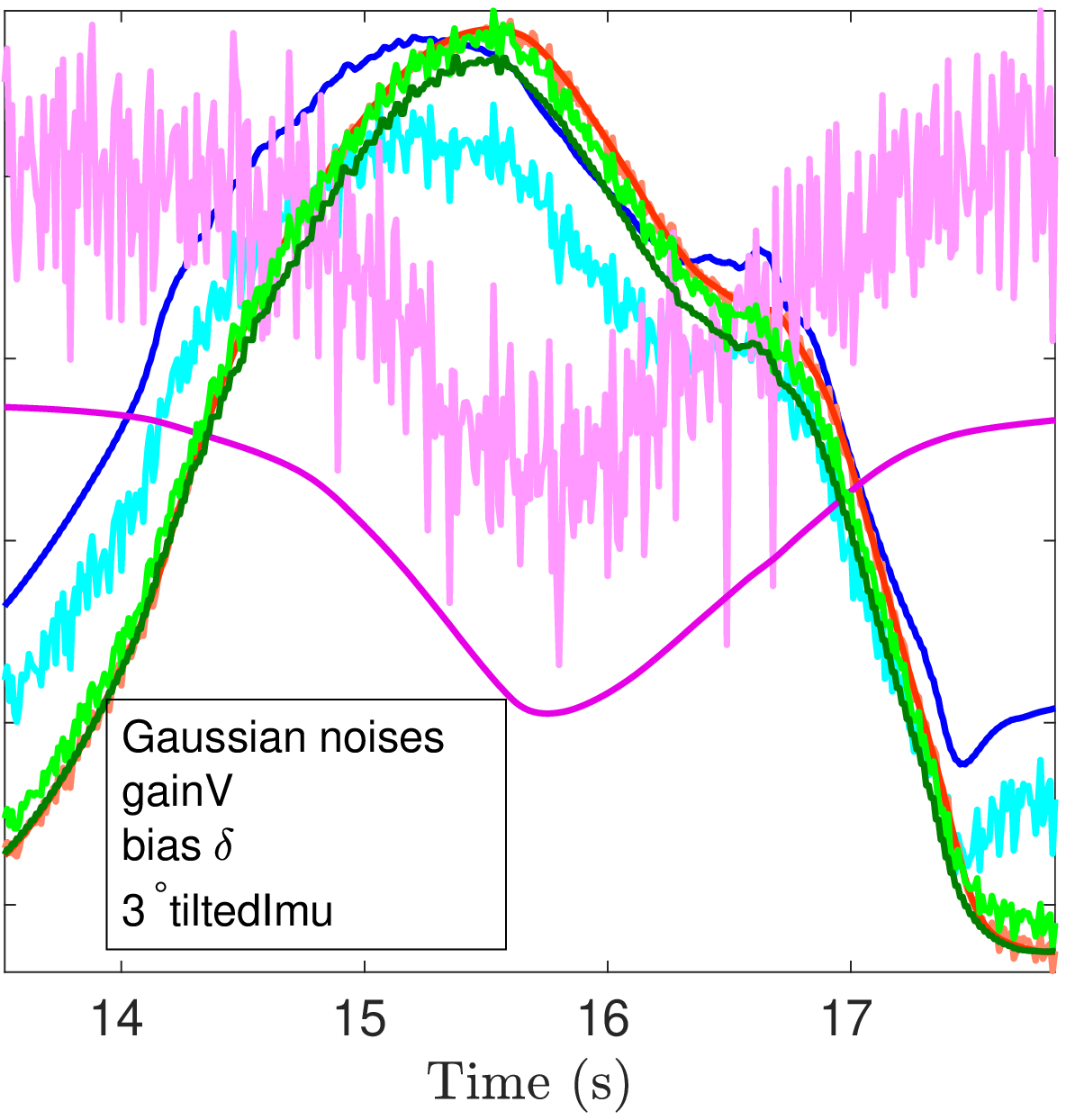}%
\label{fig:0805}}
\hfil
\subfloat[]{\includegraphics[width=0.274\textwidth]{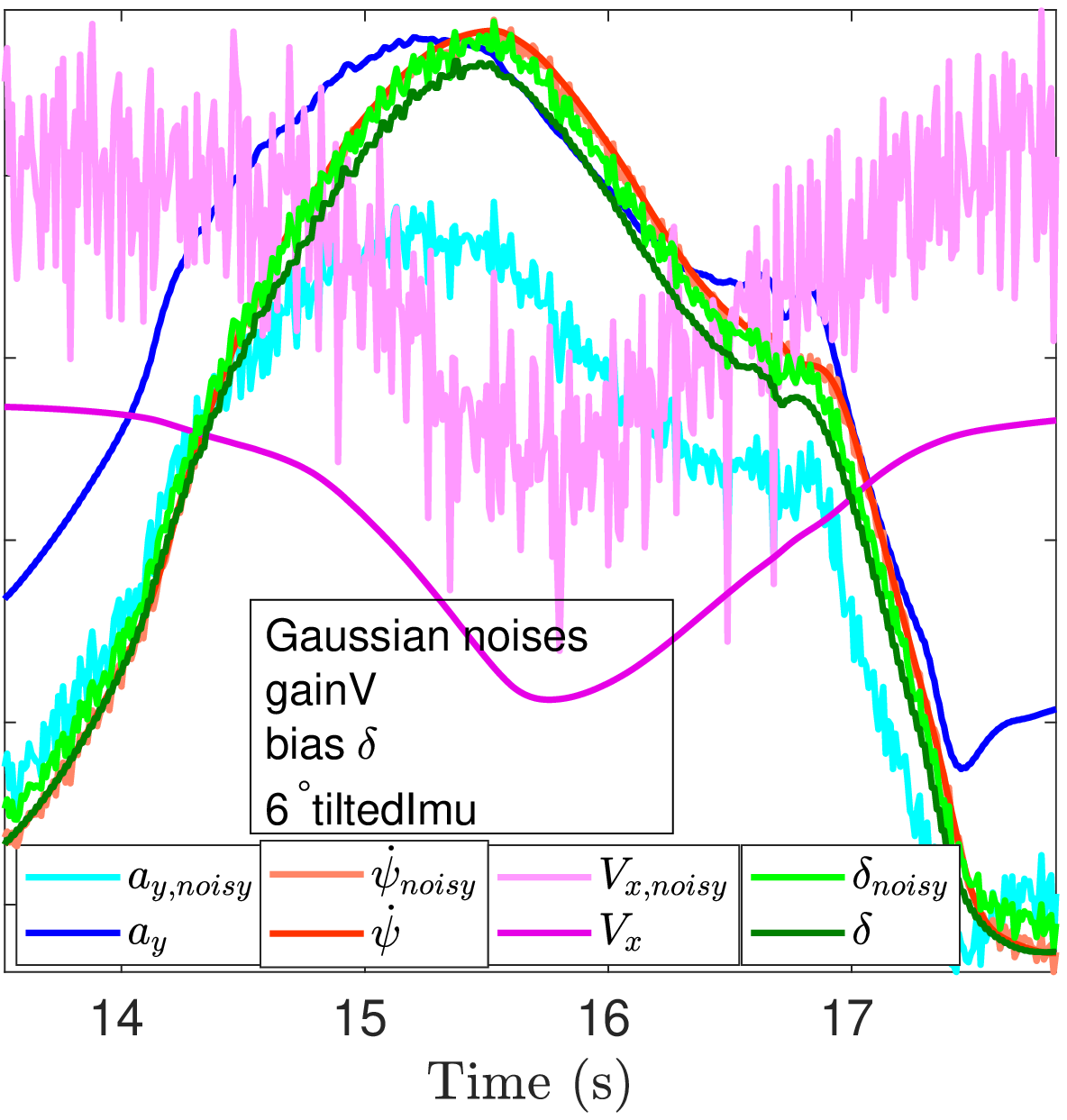}%
\label{fig:0807}}
\caption{A visual depiction of noise sets. (a) Set 1. (b) Set 2. (c) Set 4. Actual values are in dark shades and respective noise values are in light shades.
%E.g, actual $a_y$ is in dark blue and noisy $a_y$ is in light blue.
}
\label{fig:noiseSets}
\end{figure*}

\section{Experimental Setup}\label{sec4}
\noindent Figure \ref{fig:x 07_Trajectory} depicts a $438m$ long test track with eight regions labeled A-H. These regions attempt to simulate increasingly difficult maneuvers and worst environmental conditions. A is cornering with radius of $15$m ($R15$), B is cornering ($R8$) on a surface with road adherence coeffcient, $\mu=0.7$. C is double lane change, D is cornering with $\mu=0.5$, E-F are strong lateral wind with Chinese hat profile \cite{Baker2001, Baker2022}, G is a U-turn with $\mu=0.33$, and H is a slalom. The task is to follow the track center line as closely as possible, with an upper cap on vehicle speed, $V_{Ref} = 22$~km/h. It is expected that during difficult maneuvers the NMPC block modulates the vehicle speed ($V_{opt}$), to reduce the cross-track error, which is a desirable behaviour.

In order to test the limits of SRPT approach, in addition to the environmental disturbances depicted in the trajectory above, some measurement noises are also considered. They are mentioned in Table \ref{tab:worstCaseScenarios}.

For the experiments, these noise sources are grouped in the following sets:
\begin{enumerate}[label=(\roman*)]
\item Actual states, no EKF
\item Noise set 1 : EKF + Gaussian noises
\item Noise set 2 : EKF + Gaussian noises + gainV + $bias\delta$ + $3^{\circ}tiltedImu$
\item Noise set 3 : EKF + Gaussian noises + gainV + $bias\delta$ + $3^{\circ}tiltedImu + C_{F, over} + C_{R, over}$
\item Noise set 4 : EKF + Gaussian noises + gainV + $bias\delta$ + $6^{\circ}tiltedImu$
\item Noise set 5 : EKF + Gaussian noises + gainV + $bias\delta$ + $6^{\circ}tiltedImu + C_{F, over} + C_{R, over}$
\end{enumerate}

To have a visual apprehension of these noise sets, [\romannumeral 2-\romannumeral 3-\romannumeral 5] sets are depicted in figure \ref{fig:noiseSets}. These sets are recorded while the vehicle is travelling through region-B of the trajectory at $22km/h$ in the simulation framework. To further illustrate, in figure \ref{fig:0805}-\ref{fig:0807}, evident difference can be observed in noisy $a_y$, $\dot{\psi}$, and $V_x$ measurements fed to the EKF vs actual conditions.

\begin{figure}[]
\centering
\subfloat[]{\includegraphics[width=0.3\columnwidth]{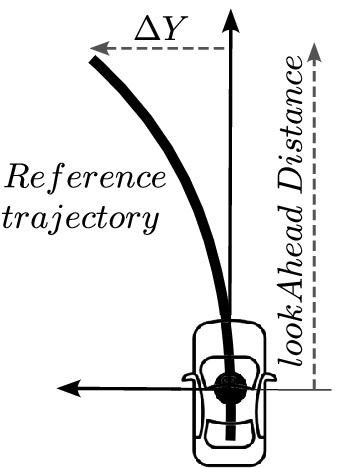}%
\label{fig:09_lookaheadDriver}}
\hfil
\subfloat[]{\includegraphics[width=0.55\columnwidth]{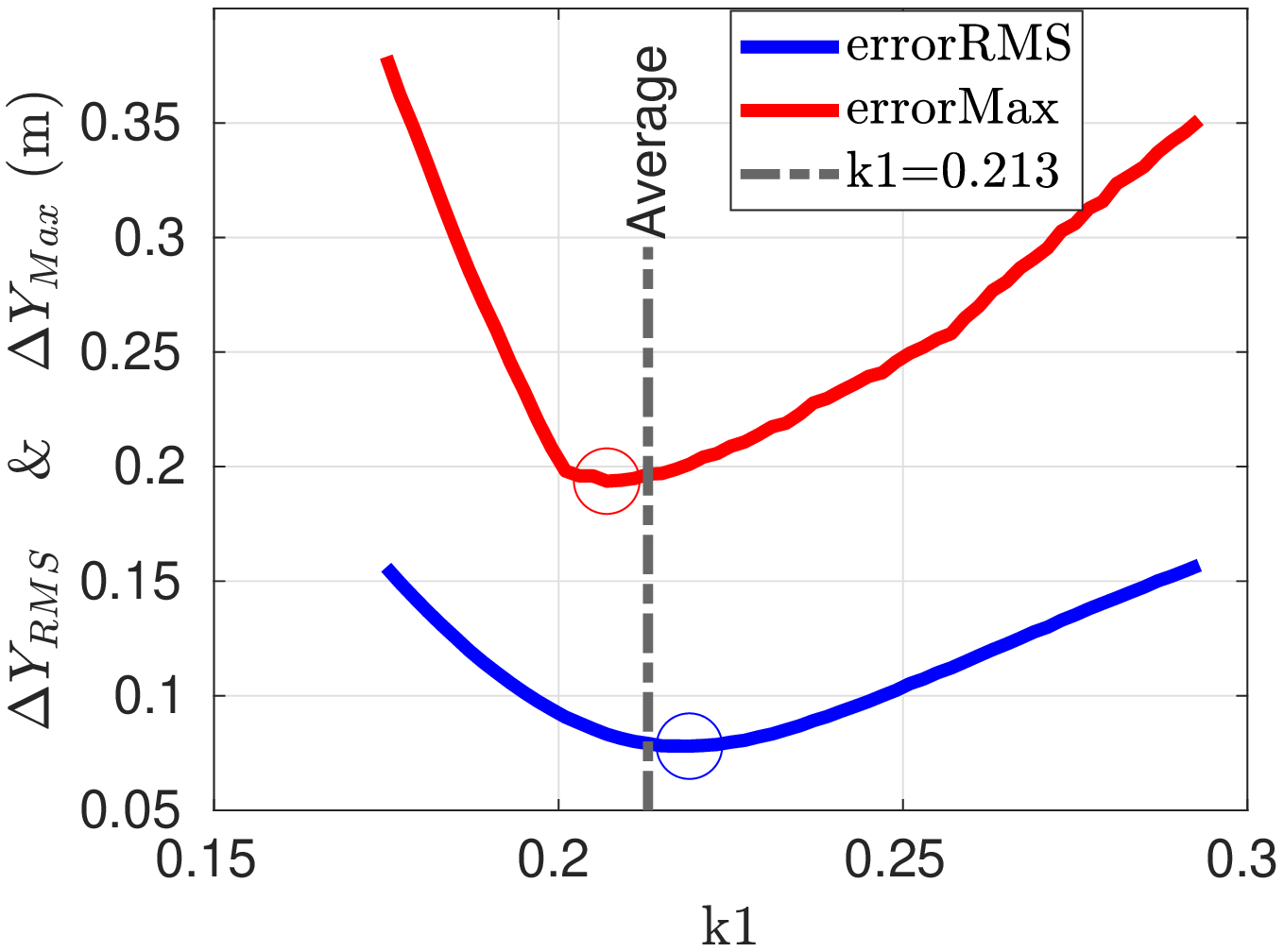}%
\label{fig:10_tunek1}}
\caption{(a) Look-ahead driver model control. (b) Tuning of $k1$ for the look-ahead driver model.}
%\label{fig:noiseSets}
\end{figure}

\subsection{Look-ahead driver model}
Performance of SRPT approach where subsequent reference poses are transmitted to vehicle is better than general vehicle teleoperation where steer commands are transmitted \cite{jai2022_3}. This is because subsequent reference poses are advance in time, this advancement in time compensates for the network delays.

For the sake of completeness of this paper, besides considering above mentioned noise cases, a separate vehicle teleoperation mode of a look-ahead driver (figure \ref{fig:09_lookaheadDriver}) is also assessed. This mode represents the general driving case, in which the human operator steers the vehicle to try to align a look-ahead point with the reference trajectory. This mode is independent of the sensor noises, as in this mode the vehicle pose is not inferred by the state estimator but rather by the human perception of the environment.

\begin{figure*}[!htb]
\centering
    \includegraphics[width=0.87\textwidth]{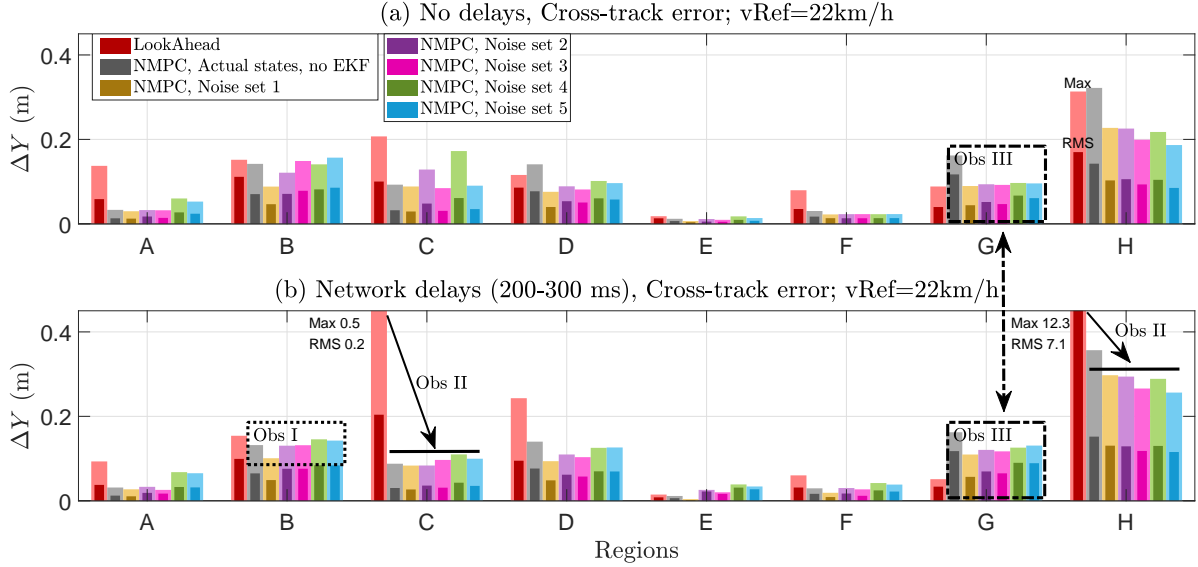}
    \caption{Region-wise Max and RMS cross-track error with various teleoperation modes at $vRef=22km/h$. (a) No network delays. (b) Under network delays (200-300ms). Consistent performances of SRPT modes in both no-delay and delay cases.}
    \label{fig:12_DeltaY}%
\end{figure*}
%Human perception may not be perfect, but to have a rigorous assessment of SRPT approach in presence of sensor noise, human perception in form of cross-track error is considered perfect.
\begin{figure}[!t]
\centering
    \includegraphics[width=.82\columnwidth]{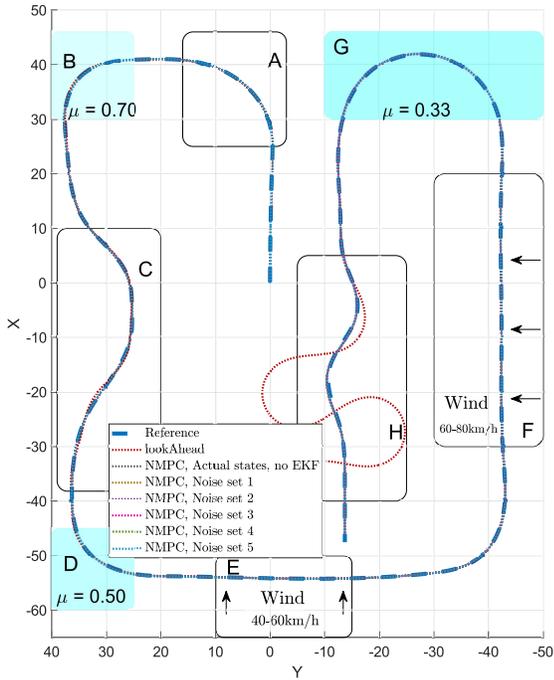}
    \caption{Trajectory traversed with various modes under network delays (200-300ms) at $vRef=22km/h$. Apart from the simple look-ahead driver mode, every other mode matches the reference trajectory}
    \label{fig:11_trajTraversed}%
\end{figure}

Look-ahead driver model based on cross-track error at the look-ahead point (motivated by \cite{Park1996}) is given by
\begin{align}
\delta &= -k_1\cdot \Delta Y \label{eq lookaheadDriverModel1}\\
lookahead Distance &=k_2\cdot V_x \label{eq lookaheadDriverModel2}\;\;\;\; .
\end{align}
\\
$\delta$ : Steer angle (rad)
\\
$k_1$ : Gain term, constant for a given vehicle longitudinal speed.
\\
$\Delta Y$ : Cross-track error of the look-ahead point from the reference trajectory.
\\
$k_2=0.90$ : look-ahead time (s).

$k_1$ and $k_2$ are considered unchanged for both no-delay and delayed teleoperation cases. Although in presence of delays a human operator can adapt his actions, but keeping $[k_1;\;k_2]$ unchanged ensures no adaptability and highlights performance deterioration due to delays.

$k_1=0.213$ is tuned to have less $\Delta Y_{RMS}$ and less $\Delta Y_{Max}$, while driving across A-B-C regions of the trajectory. Tuning is performed by iterating over a range of $k_1=[0.17:0.002:0.30]$ and observing the cross-track error (observations are shown in figure \ref{fig:10_tunek1}).
%\begin{figure}[h]
%\centering
%    \includegraphics[width=0.8\columnwidth]{Figures/10_tunek1.eps}
%    \caption{Tuning of $k1$ for the look-ahead driver model.}
%    \label{fig:10_tunek1}%
%\end{figure}

%% file: sections/05_Result.tex
\section{Results and discussion}\label{sec5}
\noindent Figure 
\ref{fig:11_trajTraversed}
, qualitatively shows better performance of SRPT approach even in the presence of all the disturbances and measurement noises. The red trajectory of look-ahead driver model resulted in big oscillations due to network delay and due to steer-rate saturation. These oscillations are not present in SRPT approach because NMPC block accounts the steer-rate limitation and consequently decelerates vehicle to allow more time to steer.

\begin{figure*}[!t]
  %\centering
  \begin{minipage}[t]{0.5\textwidth}
    \centering
    \includegraphics[width=1\textwidth]{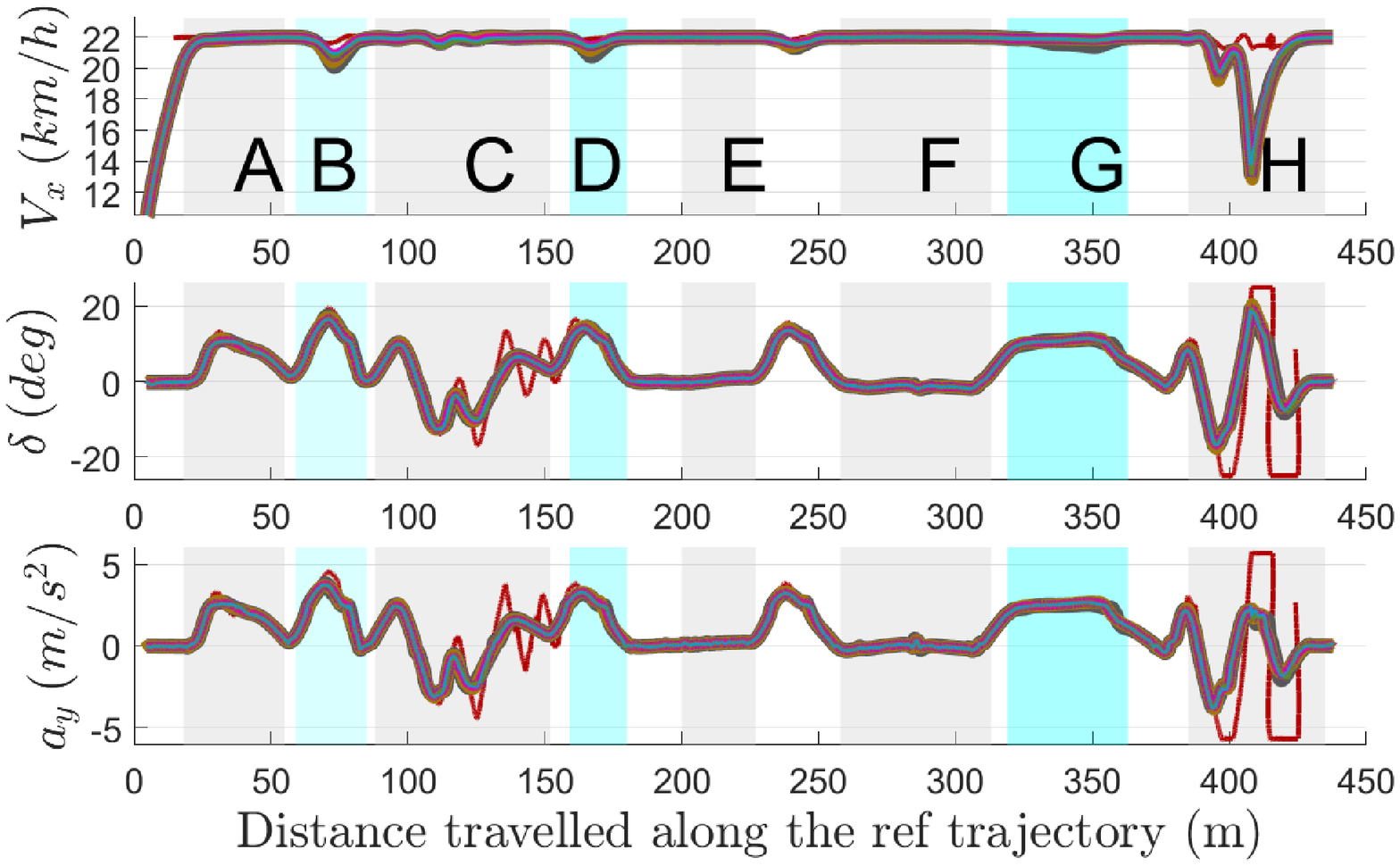}
    \caption{Vehicle speed, steer and lateral acceleration with various modes under network delays (200-300ms) at $vRef=22km/h$. Data legends are identical as shown in figure \ref{fig:12_DeltaY}. Time delay induced oscillation of vehicle steer is significantly reduced with SRPT approach even in the presence of sensor noises.}
    \label{fig:13_speedSteerAy}%
  \end{minipage}%
\hfil
  \begin{minipage}[t]{0.48\textwidth}
    \centering
    \includegraphics[width=1\textwidth]{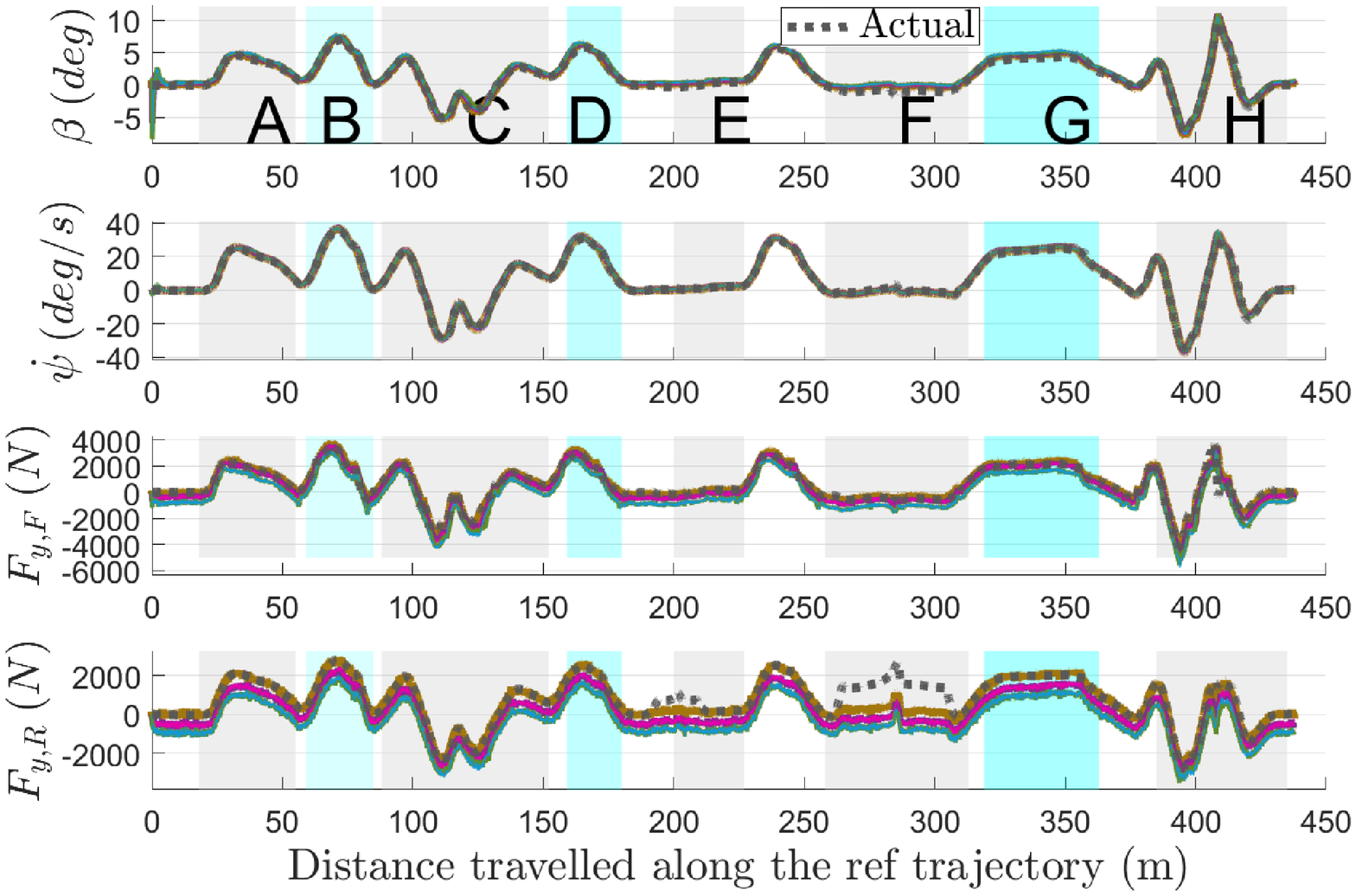}
    \caption{EKF state estimation assessment at $vRef=22km/h$. Data legends are identical as shown in figure \ref{fig:12_DeltaY}. Comparison of estimation of important states ($\beta, \dot{\psi}, F_{y, F}, F_{y, R}$) with its respective actual values.}
    \label{fig:14_nonDivergentStatesEKF}%
  \end{minipage}
\end{figure*}

Figure 
\ref{fig:12_DeltaY}
, quantitatively compares performances in terms of observed cross-track error region-wise respectively.
Larger and lighter bars are representative of the maximum values while thinner and darker bars within larger ones are representative of the rms values.
\begin{enumerate}
    \item Observation \RN{1}:  Insignificant effect of noise-sets. Across all the regions whether in no-delay or in delay cases, SRPT approach resulted consistent cross-track performance, whether true states are considered or EKF estimations are considered. 

    \item Observation \RN{2}:  Significant path-tracking performance improvement. In network delay case, for region C and region H, where maneuvers are difficult due to aggressive steer requirement, SRPT approach resulted in significant reduction in cross-track error. 

    \item Observation \RN{3}:  Relative to look-ahead driver mode, SRPT mode resulted in slightly large cross-track error in region-G (very low ground adhesion, $\mu=0.33$). This is because, the NMPC block is unaware of this big change in plant model for long time. Still, SRPT mode results similar performance in delay case as in no-delay case. E.g., \textit{Obs \RN{3}} rectangles highlight almost same cross track error in delay case as in no-delay case.
\end{enumerate}

\begin{figure*}[!b]
\centering
\subfloat[]{\includegraphics[width=1\columnwidth]{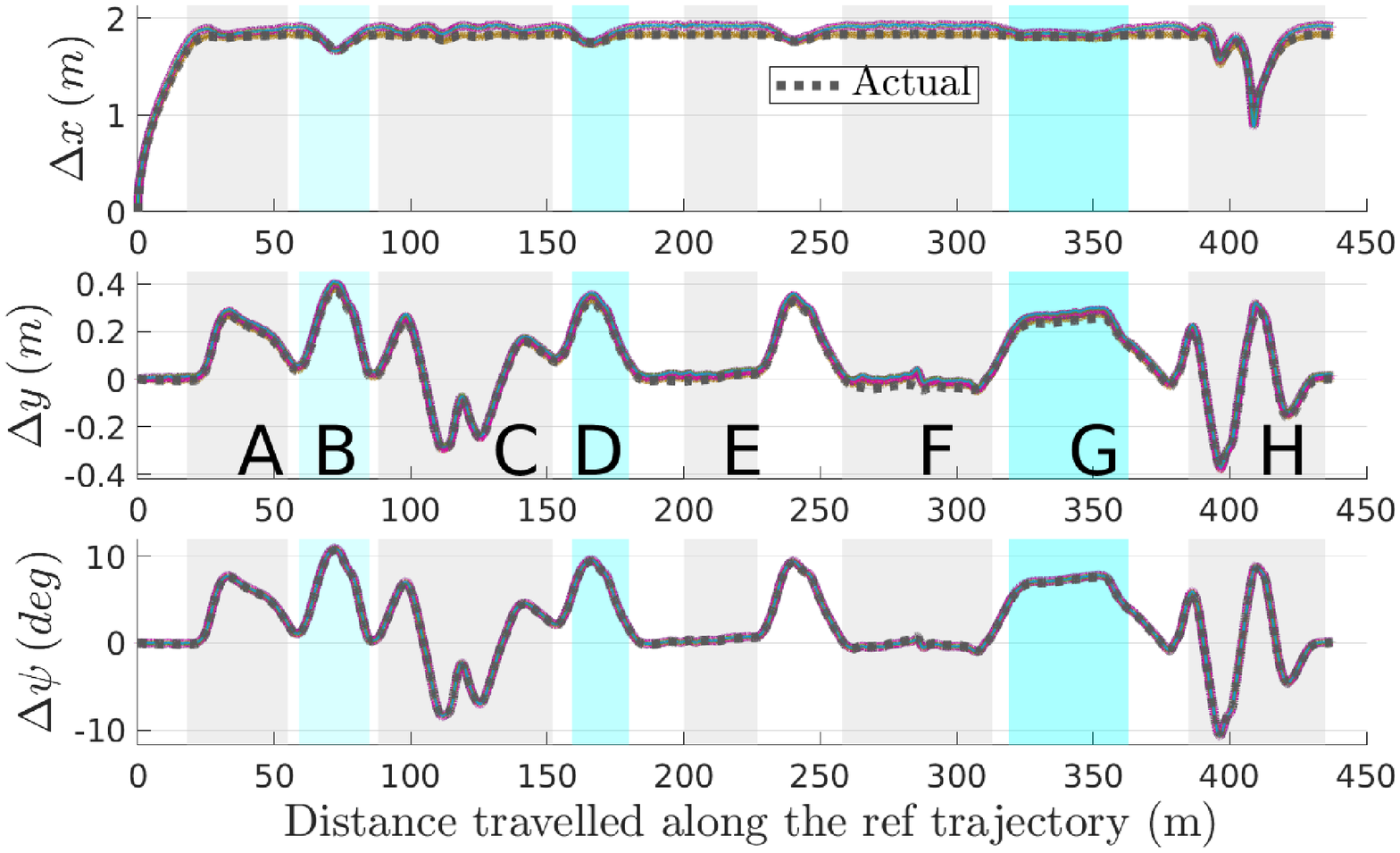}%
\label{fig:15_divergentStatesEKF}}
\hfil
\subfloat[]{\includegraphics[width=1\columnwidth]{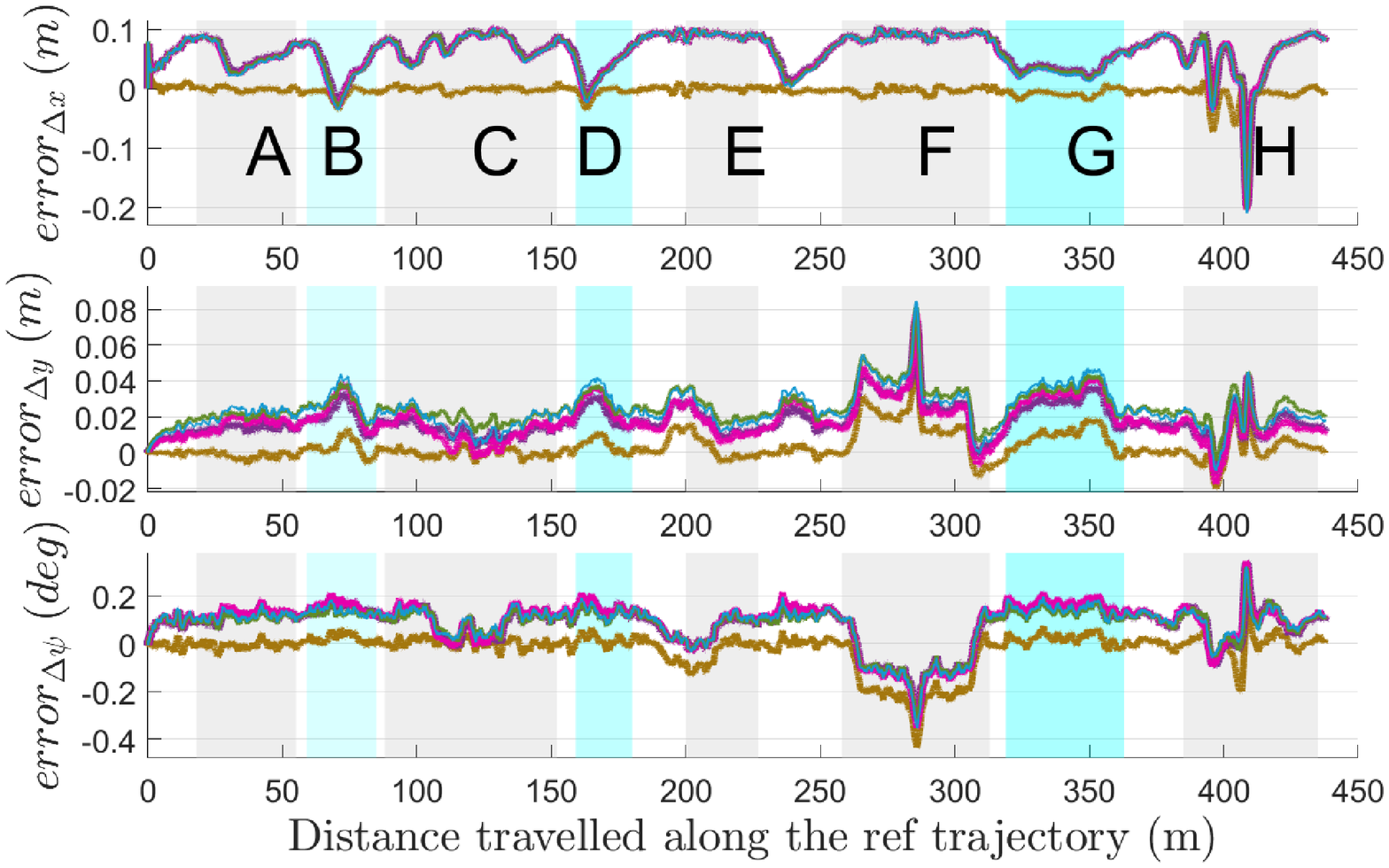}%
\label{fig:16_divergentStatesEKFerror}}
\caption{(a) EKF estimation of diverging states ($x, y, \psi$) at $vRef=22km/h$ for a time window of $\Delta=300$ms. Data legends are identical as in figure \ref{fig:12_DeltaY}. (b) EKF estimation error of diverging states ($x, y, \psi$) for the time window of $\Delta=300$ms.}
%\label{fig:noiseSets}
\end{figure*}

In figure \ref{fig:13_speedSteerAy}, modulation of vehicle speed ($V_x$) at various regions (especially in region H) of the track, results in better path tracking performance and ensures stability of the operation. With modulated vehicle speed, more time is available to steer, which eliminates oscillations in steering ($\delta$). Eventually, lateral acceleration ($a_y$) is also constrained, enhancing comfort inside the vehicle.
%% Non-divergent states

The NMPC block in figure \ref{fig:x 06_simulationPlatform} requires all the states mentioned in eq \ref{eq states}. Out of these states, $[x, y, \psi]$ states diverge because of measurement model limitation. Figure \ref{fig:14_nonDivergentStatesEKF} assess EKF state estimation accuracy by comparing estimated values of important non-diverging states $[\beta, \dot{\psi}, F_{y, F}, F_{y, R}]$ with its respective actual values. For all noise-sets states $[\beta, \dot{\psi}]$ matches quite accurately for all regions, except E and F. These are high speed wind gust regions. Even in such an extreme disturbances the estimation error is less, e.g., in region-F max error in $\beta$ estimation is around $1.5^{\circ}$. The estimation error of axle lateral forces is significant; because of the IMU tilt, some of the gravity contribution is getting added to the lateral acceleration measurement. But, overall path tracking performance with SRPT approach in vehicle teleoperation is found to be insignificantly affected by the estimation error of lateral axle forces.

\subsection{How SRPT is unaffected from diverging state estimation?}\label{sec5A}

Estimation of states $[x, y, \psi]$ diverge with time, as it is unobservable by the measurement model. Even then the SRPT approach improves vehicle teleoperation performance. 

\iffalse
\begin{figure}[b]
\centering
    \includegraphics[width=1\columnwidth]{Figures/15_divergentStatesEKF.eps}
    \caption{EKF estimation of diverging states ($x, y, \psi$) at $vRef=22km/h$ for a time window of $\Delta=300$ms. Data legends are identical as in figure \ref{fig:12_DeltaY}. For a detailed comparison, errors are presented in figure \ref{fig:16_divergentStatesEKFerror}.}%
    \label{fig:15_divergentStatesEKF}%
\end{figure}
\begin{figure}[h]
\centering
    \includegraphics[width=1\columnwidth]{Figures/16_divergentStatesEKFerror.eps}
    \caption{EKF estimation error of diverging states ($x, y, \psi$) at $vRef=22km/h$ for a time window of $\Delta=300$ms. Data legends are identical as in figure \ref{fig:12_DeltaY}.}%
    \label{fig:16_divergentStatesEKFerror}%
\end{figure}
\fi

This is due to the fact that human block in figure \ref{fig:01_mpcSchemeEkf}-figure \ref{fig:04_humanModelPerfectStates} or human model block in figure \ref{fig:05_humanModelWithEKF}-figure \ref{fig:x 06_simulationPlatform}, adds a relative pose to the received pose. The received pose is in global coordinates and the relative pose is relative to the received pose. This means, if the received pose is diverged (due to EKF state estimation), the reference-pose generated by the human block is also diverged. The NMPC block at the remote vehicle, compares the received reference-pose with the current estimation of vehicle pose, to extract the relative reference-pose. Based on that, the NMPC block optimizes for steer-speed commands. Needless to mention that, due to network delay, by the time vehicle receives a reference-pose, it has moved forward a little. What ensures the overall performance is the accuracy of vehicle pose estimation by the EKF block for a time window equal to the round-trip network delay ($200-300$ms).

To analyze the divergence in estimations of the diverging states $[x, y, \psi]$, a moving time window of $\Delta=300$ms (maximum of round-trip delay) is considered. Figure \ref{fig:15_divergentStatesEKF}, compares the estimated state evolution of states $[\Delta x, \Delta y, \Delta \psi]$ with the actual evolution of respective states for the moving time window of $300$ms (for all the noise-sets). The estimated state evolution matches the actual state evolution for all the noise-sets and in all the regions of the track.

As in this figure, errors seem unnoticeable. Figure \ref{fig:16_divergentStatesEKFerror}, highlights the error by presenting the difference between the estimated state evolution in the time window and the actual state evolution in the time window. The positive error in $x$-direction is due to the positively biased speed measurement. In the H-region the max error is $-0.2$m, this is the region where the NMPC controller decided to decelerate vehicle speed, even at this large estimation error, the SRPT approach performed better. In $y$ and $\psi$ direction, the largest error is present at the F-region, which is a region of strong cross-wind of $80$km/h. Even then, estimation error is of the order of $8$cm in lateral direction and $-0.4^{\circ}$ in heading direction. Again, the SRPT approach performed better in F-region too.

%% file: sections/07-conclusion.tex
\section{Conclusion}\label{sec6}

This article formulates an EKF-based state-estimator, that employs a minimal set of sensors, to provide estimations of vehicle state to SRPT vehicle teleoperation (previously conceptualized and assessed in \cite{jai2022_2, jai2022_3}). To test the performance of state-estimation and SRPT with state-estimation, some worst-case conditions are considered. The worst-case conditions consist of aggressive maneuvers, strong environmental disturbances, and various measurement noises. EKF state estimations are found to match actual states. Vehicle teleoperation performances of 'SRPT with actual states' and 'SRPT with estimated states' are found to be close to each other in all worst-case conditions. For sake of completeness, performance of a simple look-ahead driver is also incorporated into the analysis. SRPT teleoperation with or without EKF, performs better than the look-ahead driver model in aggressive manoeuvres, such as double-lane change and slalom. In general, SRPT modes make vehicle teleoperation unsusceptible to network delay and its variability.

Due to measurement model limitations, states that correspond to vehicle pose $[x, y, \psi]$ diverge with time, as they are unobservable. Despite this fact, SRPT with EKF performs well. This is due to the fact that the divergence of these states is negligible for a time window equal to the round-trip network delay (of $300 ms$). Hence, no exteroceptive sensor (such as GPS) is required to explicitly measure the vehicle's global pose. Being independent of GPS sensors, SRPT teleoperation becomes unaffected to problems such as unreliable GPS signals in presence of trees, tall buildings, ionospheric storms, tunnel, etc.